\RequirePackage{fix-cm}
\documentclass[smallextended]{svjour3}       
\smartqed  
\usepackage{graphicx}
\usepackage{amsmath} 
\usepackage{subfig}

\begin{document}

\title{Adaptive neural network based dynamic surface control for uncertain dual arm robots
}


\author{Dung Tien Pham \and
Thai Van Nguyen \and
Hai Xuan Le \and
Linh Nguyen \and
Nguyen Huu Thai \and
Tuan Anh Phan \and
Hai Tuan Pham \and
Anh Hoai Duong 
}


\institute{L. Nguyen \at
              Centre for Autonomous Systems, University of Technology, Sydney, New South Wales 2007, Australia \\
              Tel.: +61-2-95141225\\
              Fax: +61-2-95142655\\
              \email{vanlinh.nguyen@uts.edu.au}           
           \and
           D. T. Pham, T. V. Nguyen, H. X. Le, N. H. Thai, T. A. Phan, H. T. Pham and A. H. Duong  \at
              Department of Automatic Control, Hanoi University of Science and Technology, Hanoi 10000, Vietnam
%
%
}

\date{Received: date / Accepted: date}

\maketitle

\begin{abstract}
The paper discusses an adaptive strategy to effectively control nonlinear manipulation motions of a dual arm robot (DAR) under system uncertainties including parameter variations, actuator nonlinearities and external disturbances. It is proposed that the control scheme is first derived from the dynamic surface control (DSC) method, which allows the robot's end-effectors to robustly track the desired trajectories. Moreover, since exactly determining the DAR system's dynamics is impractical due to the system uncertainties, the uncertain system parameters are then proposed to be adaptively estimated by the use of the radial basis function network (RBFN). The adaptation mechanism is derived from the Lyapunov theory, which theoretically guarantees stability of the closed-loop control system. The effectiveness of the proposed RBFN-DSC approach is demonstrated by implementing the algorithm in a synthetic environment with realistic parameters, where the obtained results are highly promising.
\end{abstract}

\section{Introduction}	
Robots have been increasingly moving into human based environments to replace or assist human workers. More specifically, anthropomorphic or dual arm robots (DAR) have more and more played a vital role in many industrial, health care or household environments  \cite{RN51,RN52,RN53,RN54,RN55,RN78}. For instance, dual arm manipulators have been effectively employed in a diversity of tasks including assembling a car, grasping and transporting an object or nursing the elderly \cite{RN73}. In those scenarios, the DAR have been expected to behave like a human,  which is they should be able to manipulate an object similarly to what a person does \cite{RN53}. As compared to a single arm robot, the DAR have significant advantages such as more flexible movements, higher precision and greater dexterity for handling large objects \cite{RN56,RN57}. Nevertheless, since the kinematic and dynamic models of the DAR system are much more complicated than those of a single arm robot, it has more challenges to effectively and efficiently control the DAR, where synchronously coordinating the robot arms are highly expected.
	
In order to accurately and stabily track the robot arms along desired trajectories, a number of the control strategies have been proposed. For instance, the traditional methods such as nonlinear feedback control \cite{RN58} or hybrid force/position control relied on the kinematics and statics \cite{RN59,RN60} have been proposed to simultaneously control both of the arms. In the works \cite{RN61,RN62,RN63}, the authors have proposed to utilize the impedance control by considering the dynamic interaction between the robot and its surrounding environment while guaranteeing the desired movements. More importantly, robustness of the control performance is also highly prioritized in consideration of designing a controller for a highly uncertain and nonlinear DAR system. In literature of the modern control theory, sliding mode control (SMC) demonstrates a diverse ability to robustly control any system. Since the pioneer work  \cite{RN64}, the variable structure SMC has enjoyed widespread use and attention in many applications \cite{RN10,RN11,RN12,RN65,RN67,RN68,RN77,Le2019b,Le2017,VA2018,Le2019,Le2019a}. Nonetheless, due to presence of discontinuities, the SMC law may cause undesirable oscillations, which is also called the chatterring phenomenon. Park \textit{et al.} \cite{RN13} proposed a saturation function to replace the discontinuous signum function in the control signal to reduce effects of the chattering. Recently, the chattering phenomenon can be eliminated by designing the controller without a discontinuous term \cite{Le2017,VA2018}. For robustly controlling nonlinear systems with unmatched uncertainties, the SMC control law is usually designed in conjunction with the backstepping method, where the sliding surface is aggregated in the last step \cite{RN14,RN17}. For instance, Chen \textit{et al.} \cite{RN75} developed a backstepping sliding mode controller to enhance the global ultimate asymptotic stability and invariability to uncertainties in a nonholonomic wheeled mobile manipulator. To address the problem of explosion of terms associated with the integrator backstepping technique, Swaroop \textit{et al.} \cite{RN19} proposed the dynamic surface control (DSC) method by using a first-order low-pass filter in the synthetic input.
	
Nonetheless, the aforesaid traditional control techniques are not really practical when they require to accurately model all the nonlinear dynamics of the DAR system, where its unknown parameters are highly uncertain and not easily estimated. It is noted that uncertainties of the DAR system can practically lead to degradation of its control performance. Furthermore, a number of unexpected disturbances and obstacles in the working environments can cause the DAR system to be unstable. To address the issues of accurately modelling all the nonlinear dynamics and estimating the unknown and uncertain parameters, some modern control approaches based on fuzzy logic or artificial neural network have been proposed in the past decades. For instance, by the use of the adaptive learning and function approximations, Lee and Choi \cite{RN70} introduced a radial basis function network (RBFN) for approximating the nonlinear dynamics of a SCARA-type robot manipulator. Similarly, Wang \textit{et al.} \cite{RN72} employed the approximation of a neural network to deal with the nonlinearities and uncertainties of a single robot manipulator, where errors caused by the neural network approximation can be estimated by a proposed control robust term. In addition, the authors in \cite{RN73} designed an adaptive control system for a humanoid robot by using the RBFN to develop a scheme to adaptively estimate unknown and uncertain dynamics of the robot. Based on a multi-input multi-output fuzzy logic unit, Jiang \textit{et al.} \cite{RN74} proposed an algorithm to adaptively estimate the dynamics of the DAM, given its nonsysmmetric deadzone nonlinearity. In the context of adaptive DSC, it was proposed to employ fuzzy techniques and neural networks to adaptively estimate parameters for the control laws in uncertain nonlinear systems \cite{Luo2009} and nonlinear systems with uncertain time delays \cite{Wu2016}, respectively.

In this paper, we propose an adaptive control strategy based on the DSC method and the RBFN to effectively and efficiently control the DAR system. The proposed approach provides the DAR system not only adaptive estimation of the nonlinear dynamics but also robustness to system uncertainties including the system parameter variations, actuator nonlinearities and external disturbances. In other words, the aggregated control scheme based on the DSC technique enables the manipulators to be capable of efficiently tracking the desired trajectories given large variation of the system information such as the undetermined volume and mass of the payload and significantly reducing chattering influences. The RBFN allows the proposed controller to be able to adaptively estimate the nonlinear and uncertain parameters of the DAR system. More importantly, the adaptation mechanism is designed based on Lyapunov method, which mathematically guarantees the stability of the closed-loop control system. The proposed algorithm was extensively validated in the synthetic environments, where the obtained results are highly promising.

	The rest of the paper is arranged as follows. We first introduce a model of the DAR system in Section \ref{sec_2}. We then present how to construct a RBFN-DSC controller for the DAR system based on the DSC and RBFN in Section \ref{sec_3}. Section \ref{sec_4} discusses validation of the proposed approach in a simulation environment before conclusions are drawn in Section \ref{sec_5}.
	
\section{Dual Arm Robot Model}
\label{sec_2}
\begin{figure}[t]
		\centering
		\includegraphics[width=0.8\linewidth]{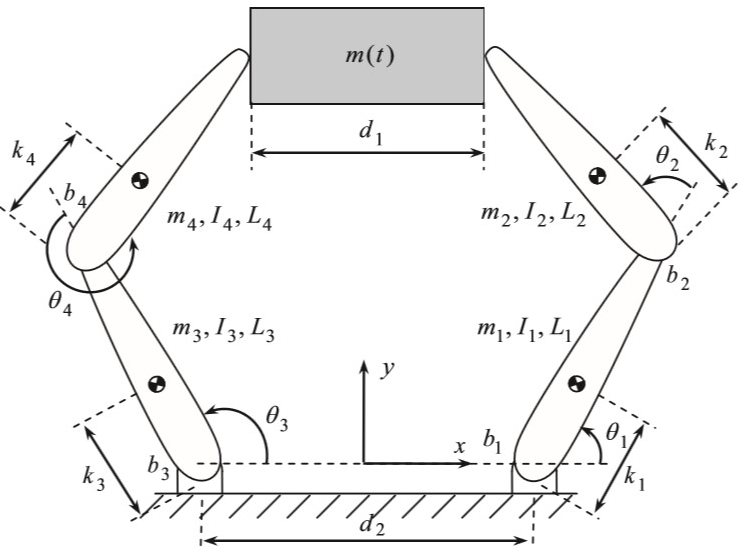}
		\caption{Dual arm robot modelling}
		\label{fig1}
	\end{figure}

Lets consider a dual two degree of free (DoF) arm robot that cooperatively manipulates an object with mass of $m$ as pictorically shown in Fig. \ref{fig1}. It is assumed that both the manipulators rigidly attach to the load so that there is no slip between the grasping points and the grasped load. Let $m_i, I_i, l_i$ denote the mass, mass moment of inertia and length of the corresponding link in the model, respectively. We also define $d_1$ and $d_2$ as the length of the object and distance between the two arms at the robot's base. The distance from the mass centre of a link to a joint is denoted as $k_i$ while the joint angle between a link and the base or its preceding link is denoted as $\theta_i$. 
	
	\begin{figure}[t]
		\centering
		\includegraphics[width=0.8\linewidth]{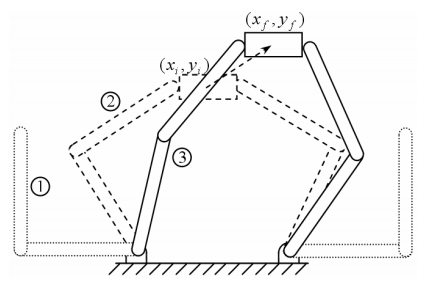}
		\caption{Operational motions of dual arm robot}
		\label{fig2}
	\end{figure}

Operationally, in this work we consider that the robot manipulators make motions on the horizontal $xy$ plane. In other words, the robot arms first move towards the object. After the manipulators are firmly attached to the load, the robot then picks the object up and transports it to a new position by adjusting the motions to robustly follow the given trajectory, demonstrated in Fig. \ref{fig2}, where ($x_i$,$y_i$) and ($x_f$,$y_f$) are the initial and final locations of the payload, respectively. We let $x_m$ and $y_m$ denote the mass center of the payload on the $xy$ plane, the trajectory of the object can be specified by
\begin{equation}
\begin{array}{r@{}l@{\qquad}l} 
	{{x}_{m}}&{}=\frac{{{d}_{2}}}{2}+{{l}_{1}}\cos {{\theta}_{1}}+{{l}_{2}}\cos ({{\theta}_{1}}+{{\theta}_{2}})-\frac{{{d}_{1}}}{2} \\ 
	&{}=-\frac{{{d}_{2}}}{2}+{{l}_{3}}\cos {{\theta}_{3}}+{{l}_{4}}\cos ({{\theta}_{3}}+{{\theta}_{4}})+\frac{{{d}_{1}}}{2}, \\ \nonumber
	{{y}_{m}}&{}={{l}_{1}}\sin {{\theta}_{1}}+{{l}_{2}}\sin ({{\theta}_{1}}+{{\theta}_{2}}) \\ \nonumber
	&{}={{l}_{3}}\sin {{\theta}_{3}}+{{l}_{4}}\sin ({{\theta}_{3}}+{{\theta}_{4}}). 
	\end{array} 
\end{equation}

\begin{figure}[t]
		\centering
		\includegraphics[width=0.8\linewidth]{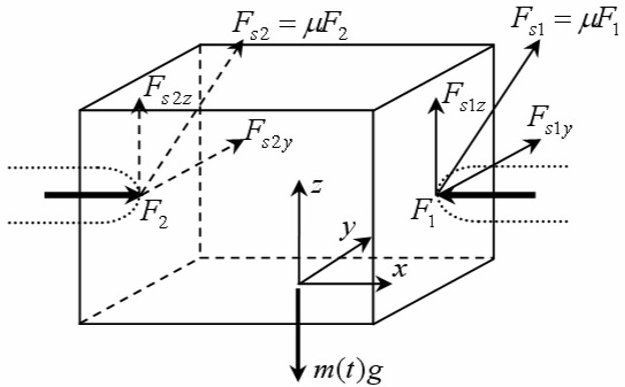}
		\caption{Physical model of the robot arms}
		\label{fig3}
	\end{figure}

In order to transport the object to a new position, the robot manipulators apply forces $F_1$ and $F_2$ to the payload as illustrated in Fig. \ref{fig3}. On the other hands, to rigidly hold the load up, friction forces $F_{s1}$ and $F_{s2}$ are needed. Let $F_{siy}$ and $F_{siz}$ denote the components of the friction forces in $y$ and $z$ directions, respectively. To prevent the load from rotating around $y$ and $z$ axes, it is supposed that $F_{s1y}=F_{s2y}$ and $F_{s1z}=F_{s2z}$. Then the dynamic equations of the object are as follows,
	\begin{equation}\label{eq3}
	\begin{array}{r@{}l@{\qquad}l}
	&m{{\ddot{x}}_{m}}={{F}_{2}}-{{F}_{1}}, \\
	&m{{\ddot{y}}_{m}}=2{{F}_{s1y}}=2{{F}_{s2y}}, \\
	&mg=2{{F}_{s1z}}=2{{F}_{s2z}},
	\end{array}
	\end{equation}
where 
\begin{align}
	\begin{array}{r@{}l@{\qquad}l}
	{{\ddot{x}}_{m(t)}}=&-{{L}_{1}}\left( {{{\dot{\theta }}}_{1}}\cos {{\theta }_{1}}+{{{\ddot{\theta }}}_{1}}\sin {{\theta }_{1}} \right)\\&-{{L}_{2}}\left[ {{\left( {{{\dot{\theta }}}_{1}}+{{{\dot{\theta }}}_{2}} \right)}^{2}}\cos \left( {{\theta }_{1}}+{{\theta }_{2}} \right)\right]\\
	&-{{L}_{2}}\left[{{\left( {{{\ddot{\theta }}}_{1}}+{{{\ddot{\theta }}}_{2}} \right)}^{2}}\sin \left( {{\theta }_{1}}+{{\theta }_{2}} \right) \right],
	\end{array}
	\end{align}
	\begin{align}
	\begin{array}{r@{}l@{\qquad}l}
	{{\ddot{y}}_{m(t)}}=&{{L}_{3}}\left( {{{\ddot{\theta }}}_{3}}\cos {{\theta }_{3}}+{{{\dot{\theta }}}_{3}}^{2}\sin {{\theta }_{3}} \right)\\&-{{L}_{4}}\left[ \left( {{{\ddot{\theta }}}_{3}}+{{{\ddot{\theta }}}_{4}} \right)\cos \left( {{\theta }_{3}}+{{\theta }_{4}} \right)\right]\\
	&+{{L}_{4}}\left[{{\left( {{{\dot{\theta }}}_{3}}+{{{\dot{\theta }}}_{4}} \right)}^{2}}\sin \left( {{\theta }_{3}}+{{\theta }_{4}} \right) \right],
	\end{array}
	\end{align}
and $g=9.8 m/s^2$. And the relationship between the applied forces and the friction forces is presented by	
	\begin{equation}\label{eq6}
	\begin{array}{r@{}l@{\qquad}l}
	{{F}_{s1y}}^{2}+{{(\frac{mg}{2})}^{2}}<{{(\mu {{F}_{1}})}^{2}}, \\
	{{F}_{s2y}}^{2}+{{(\frac{mg}{2})}^{2}}<{{(\mu {{F}_{2}})}^{2}},
	\end{array}
	\end{equation}
where $\mu$ is the friction coefficient in dry condition.

If ${{\ddot{x}}_{m(t)}}\ge 0$, both the applied forces $F_1$ and $F_2$ can be computed by
	\begin{equation}\label{eq8}
	\begin{array}{r@{}l@{\qquad}l}
	{{F}_{1}}&=\frac{1}{\mu }\sqrt{{{\left( \frac{m{{{\ddot{y}}}_{m}}}{2} \right)}^{2}}+{{\left( \frac{mg}{2} \right)}^{2}}}, \\
	{{F}_{2}}&=\frac{1}{\mu }\sqrt{{{\left( \frac{m{{{\ddot{y}}}_{m}}}{2} \right)}^{2}}+{{\left( \frac{mg}{2} \right)}^{2}}}+m{{\ddot{x}}_{m}}.
	\end{array}
	\end{equation}
Nonetheless, if ${{\ddot{x}}_{m(t)}}<0$, those forces can be obtained by
	\begin{equation}\label{eq10}
	\begin{array}{r@{}l@{\qquad}l}
	{{F}_{1}}&=\frac{1}{\mu }\sqrt{{{\left( \frac{m{{{\ddot{y}}}_{m}}}{2} \right)}^{2}}+{{\left( \frac{mg}{2} \right)}^{2}}}-m{{\ddot{x}}_{m}}, \\
	{{F}_{2}}&=\frac{1}{\mu }\sqrt{{{\left( \frac{m{{{\ddot{y}}}_{m}}}{2} \right)}^{2}}+{{\left( \frac{mg}{2} \right)}^{2}}}.
	\end{array}
	\end{equation}
	
	\

By the use of Lagrange multipliers, the dynamic model of the dual arm robot manipulating the payload can be summarized as follows,
	\begin{equation}\label{eq12}
	M(\theta)\ddot{\theta}+C(\theta,\dot{\theta})\dot{\theta}={u}+{{J}^{T}}(\theta)F(\theta,\dot{\theta},\ddot{\theta})-{{T}_{d}}-\beta, \nonumber
	\end{equation}
where $u$ is a $4\times1$ control torque input vector, ${{T}_{d}}$ is a $4\times1$ vector presenting the noise effects on the robot arms and $\beta$ denotes the viscous friction forces on all the joints, which are specified as follows,
\begin{equation}
\begin{array}{r@{}l@{\qquad}l}
	\theta&={{\left[ \begin{matrix}
			{{\theta}_{1}}\; {{\theta}_{2}}\; {{\theta}_{3}}\; {{\theta}_{4}}   \nonumber
			\end{matrix} \right]}^{T}}, \\ \nonumber
	u&={{\left[ \begin{matrix}
			{{u}_{1}} \; {{u}_{2}} \; {{u}_{3}} \; {{u}_{4}}  \nonumber
			\end{matrix} \right]}^{T}}, \\ \nonumber
	F&={{\left[ \begin{matrix}
			{{F}_{1}} \; {{F}_{s1y}} \; {{F}_{2}} \; {{F}_{s2y}}   \nonumber
			\end{matrix} \right]}^{T}}, \\ \nonumber
	{{T}_{d}}&={{\left[ \begin{matrix}
			{{T}_{d1}} \;{{T}_{d2}} \; {{T}_{d3}}\; {{T}_{d4}}   \nonumber
			\end{matrix} \right]}^{T}}, \\ \nonumber
	\beta &={{\left[ \begin{matrix}
			{{b}_{1}}{{{\dot{\theta}}}_{1}} \; {{b}_{2}}{{{\dot{\theta}}}_{2}} \; {{b}_{3}}{{{\dot{\theta}}}_{3}}\; {{b}_{4}}{{{\dot{\theta}}}_{4}} 
			\end{matrix} \right]}^{T}},  \nonumber
			\end{array}
	\end{equation} 
where $b_i$ is the viscous friction at the $i^{th}$ joint.
$M(\theta)$ is a $4\times4$ matrix of the mass moment of inertia, whose components are specified by
\begin{equation}
\begin{array}{r@{}l@{\qquad}l}
	& {{m}_{11}}={{A}_{1}}+{{A}_{2}}+2{{A}_{3}}\cos {{\theta}_{2}}, \\ \nonumber 
	& {{m}_{12}}={{m}_{21}}={{A}_{2}}+{{A}_{3}}\cos {{\theta}_{2}}, \\ \nonumber 
	& {{m}_{22}}={{A}_{2}}, \\ \nonumber
	& {{m}_{13}}={{m}_{14}}={{m}_{23}}={{m}_{24}}=0, \\ \nonumber
	& {{m}_{33}}={{A}_{4}}+{{A}_{5}}+2{{A}_{6}}\cos {{\theta}_{4}}, \\ \nonumber 
	& {{m}_{34}}={{m}_{43}}={{A}_{5}}+{{A}_{6}}\cos {{\theta}_{4}},\\ \nonumber 
	& {{m}_{44}}={{A}_{5}}, \\ \nonumber
	& {{m}_{31}}={{m}_{32}}={{m}_{41}}={{m}_{42}}=0 \nonumber
	\end{array}
\end{equation}
with
\begin{equation}
\begin{array}{r@{}l@{\qquad}l}
	& {{A}_{1}}={{m}_{1}}k_{1}^{2}+{{m}_{2}}l_{1}^{2}+{{I}_{1}}, \\ 
	& {{A}_{2}}={{m}_{2}}k_{2}^{2}+{{I}_{2}}, \\
	& {{A}_{3}}={{m}_{2}}{{l}_{1}}{{k}_{2}}, \\ 
	& {{A}_{4}}={{m}_{3}}k_{3}^{2}+{{m}_{4}}l_{3}^{2}+{{I}_{3}}, \\ 
	& {{A}_{5}}={{m}_{4}}{{k}_{4}}^{2}+{{I}_{4}}, \\
	& {{A}_{6}}={{m}_{4}}{{l}_{3}}{{k}_{4}}. 
	\end{array}
	\end{equation}
$C(\theta,\dot{\theta})$ is a $4\times1$ Coriolis-centripetal vector, whose elements are computed by
\begin{equation}
\begin{array}{r@{}l@{\qquad}l}
	& {{c}_{11}}=-{{A}_{3}}\sin {{\theta}_{2}}(\dot{\theta}_{2}^{2}+{{{\dot{\theta}}}_{1}}{{{\dot{\theta}}}_{2}})+{{b}_{1}}{{{\dot{\theta}}}_{1}}, \\ 
	& {{c}_{21}}={{A}_{3}}\dot{\theta}_{1}^{2}\sin {{\theta}_{2}}+{{b}_{2}}{{{\dot{\theta}}}_{2}}, \\ 
	& {{c}_{31}}=-{{A}_{6}}\sin {{\theta}_{4}}(\dot{\theta}_{4}^{2}+{{{\dot{\theta}}}_{3}}{{{\dot{\theta}}}_{4}})+{{b}_{3}}{{{\dot{\theta}}}_{3}}, \\ 
	& {{c}_{41}}={{A}_{6}}\dot{\theta}_{3}^{2}\sin {{\theta}_{4}}+{{b}_{2}}{{{\dot{\theta}}}_{4}}. 
	\end{array}
	\end{equation}
Furthermore, $J$ is a $4\times4$ Jacobian matrix with the elements obtained by
\begin{equation}
\begin{array}{r@{}l@{\qquad}l}
	& {{J}_{11}}=-{{L}_{1}}\sin {{\theta}_{1}}-{{L}_{2}}\sin ({{\theta}_{1}}+{{\theta}_{2}}), \\ 
	& {{J}_{12}}=-{{L}_{1}}\cos {{\theta}_{1}}-{{L}_{2}}\cos ({{\theta}_{1}}+{{\theta}_{2}}), \\ 
	& {{J}_{13}}={{J}_{14}}=0, \\ 
	& {{J}_{21}}=-{{L}_{2}}\sin ({{\theta}_{1}}+{{\theta}_{2}}), \\
	& {{J}_{22}}=-{{L}_{2}}\cos ({{\theta}_{1}}+{{\theta}_{2}}), \\
	& {{J}_{23}}={{J}_{24}}=0, \\ 
	& {{J}_{31}}={{J}_{32}}=0, \\
	& {{J}_{33}}={{L}_{3}}\sin {{\theta}_{3}}+{{L}_{4}}\sin ({{\theta}_{3}}+{{\theta}_{4}}), \\ 
	& {{J}_{34}}=-{{L}_{3}}\cos {{\theta}_{3}}-{{L}_{4}}\cos ({{\theta}_{3}}+{{\theta}_{4}}), \\ 
	& {{J}_{41}}={{J}_{42}}=0, \\ 
	& {{J}_{43}}={{L}_{4}}\sin ({{\theta}_{3}}+{{\theta}_{4}}), \\ 
	& {{J}_{44}}=-{{L}_{4}}\cos ({{\theta}_{3}}+{{\theta}_{4}}).
	\end{array}
	\end{equation}

\section{Control Approach}
\label{sec_3}
In order to design a control law to efficiently and automatically adjust the robot manipulators, we first discuss a control scheme based on the dynamic surface control method. It is noted that due to the system uncertainties including parameter variations, actuator nonlinearities and external disturbances, system parameters in the designed controller are practically uncertain and unknown; then we introduce a radial basis function network based technique to adaptively estimate those uncertain and unknown dynamics.

Generally speaking, the dynamic model of the dual arm robot (DAR) (\ref{eq12}) can be represented as follows,
	\begin{equation}\label{eq14}
\begin{array}{r@{}l@{\qquad}l}
	& {{{{\dot{x}}}}_{1}}={{{{x}}}_{2}} \\ 
& {{{{\dot{x}}}}_{2}}={{M}^{-1}}(\theta )u+{{M}^{-1}}(\theta )[{{J}^{T}}\left( \theta  \right)F(\theta ,\dot{\theta },\ddot{\theta })-T_d-\beta-C(\theta ,\dot{\theta })],
	\end{array}
	\end{equation}
where ${{{x}}_{1}}={{({{\theta }_{1}},{{\theta }_{2}},{{\theta }_{3}},{{\theta }_{4}})}^{T}}$ and  ${{{x}}_{2}}={{({{\dot{\theta }}_{1}},{{\dot{\theta }}_{2}},{{\dot{\theta }}_{3}},{{\dot{\theta }}_{4}})}^{T}}$.
Let $K(\theta,\dot{\theta},\ddot{\theta})={{J}^{T}}(\theta)F(\theta,\dot{\theta},\ddot{\theta})-C(\theta,\dot{\theta})-G(\theta)-\beta -{{T}_{d}}$, the (\ref{eq14}) can be simplified by
\begin{equation}\label{eq15}
\begin{array}{r@{}l@{\qquad}l}
& {{{{\dot{x}}}}_{1}}={{{{x}}}_{2}} \\ 
& {{{{\dot{x}}}}_{2}}={{M}^{-1}}(\theta ){u}+{{M}^{-1}}(\theta ).K(\theta ,\dot{\theta },\ddot{\theta }) \\ 
\end{array}.
\end{equation}
It can be clearly seen that the system uncertainties are now incorporated into $K(\theta,\dot{\theta},\ddot{\theta})$ that presents the complex nonlinear dynamic of the robot. For the purpose of simplicity, $K(\theta,\dot{\theta},\ddot{\theta})$ and $K$ will be used interchangeably.
	
\subsection{Dynamic surface control for certain DAR systems}
\label{sec_3a}
The aim of controlling a dual arm robot is to guarantee that ${{{x}}_{1}}$ tracks the reference ${{{x}}_{1r}}$. Therefore, we propose to design a control law using the dynamic surface control (DSC) structure as sequentially expressed by the following steps.

\textbf{Step 1}:
Let
\begin{equation}\label{eq16}
{{{z}}_{1}}={{{x}}_{1}}-{{{x}}_{1r}}
\end{equation}
denote the vector of tracking errors, and consider the first Lyapunov function candidate as follows,
\begin{equation}\label{eq17}
{{V}_{1}}=\frac{1}{2}{{{z}}_{1}}^{T}{{{z}}_{1}}.
\end{equation}
If differentiating ${{V}_{1}}$ with respect to time, one obtains
\begin{equation}\label{eq18}
\begin{array}{r@{}l@{\qquad}l}
{{\dot{V}}_{1}}&={{{z}}_{1}}^{T}{{\dot{{z}}}_{1}}={{{z}}_{1}}^{T}\left( {{{{\dot{x}}}}_{1}}-{{{{\dot{x}}}}_{1r}} \right)={{{z}}_{1}}^{T}({{{x}}_{2}}-{{{\dot{x}}}_{1r}})\\
&=-{{c}_{1}}{{{z}}_{1}}^{T}{{{z}}_{1}}+{{{z}}_{1}}^{T}\left( {{{{x}}}_{2}}-{{{{\dot{x}}}}_{1r}}+{{c}_{1}}{{{{z}}}_{1}} \right),
\end{array}
\end{equation}
where $c_1$ is a positive definite diagonal matrix.

\textbf{Step 2}:
Let
\begin{equation}\label{eq19}
{{{z}}_{2}}={{{x}}_{2}}-{{{\alpha }}_{2f}}
\end{equation}
define the error between the input ${{{x}}_{2}}$ and the virtual control ${{{\alpha }}_{2f}}$, which is also an output of the first-order filter when putting ${\alpha }$ through. ${\alpha}$ is assumed a virtual control, given by
\begin{equation}\label{eq23}
{\alpha }={{\dot{x}}}_{1r}-{{c}_{1}}{{{z}}_{1}}. 
\end{equation}
If the first-order low-pass filter \cite{RN19} is presented by 
\begin{equation}\label{eq20}
\tau {{{\dot{\alpha }}}_{2f}}+{{{\alpha }}_{2f}}={\alpha },
\end{equation}
where ${{\alpha }_{2f}}(0)=\alpha (0)$ and $\tau >0$ is the filter constant. Then, the sliding surface is defined as follow,
\begin{equation}\label{eq21}
{s}={{\lambda }}{{{z}}_{1}}+{{{z}}_{2}},
\end{equation}
where $\lambda $ is a positive definite diagonal matrix. The derivative of the sliding surface can be easily obtained by \cite{RN14}
\begin{equation}\label{eq22}
\begin{array}{r@{}l@{\qquad}l}
 \dot{s}&={{\lambda }}{{{\dot{{z}}}}_{1}}+{{{{\dot{z}}}}_{2}} \\ 
&={{\lambda }}{{{\dot{{z}}}}_{1}}+({{{\dot{{x}}}_{2}}}-{{{{\dot{\alpha }}}}_{2f}}) \\ 
 &={{\lambda }}{{{\dot{{z}}}}_{1}}+{{M}^{-1}}K+{{M}^{-1}}{u}-{{{{\dot{\alpha }}}}_{2f}} \\ 
 &={{\lambda }}{{{\dot{{z}}}}_{1}}+{{M}^{-1}}\left( K+{u}-M{{{{\dot{\alpha }}}}_{2f}} \right).
\end{array}
\end{equation}

Now, let's consider the second Lyapunov function candicate as follows,
\begin{equation}\label{eq24}
{{V}_{2}}={{V}_{1}}+\frac{1}{2}{{{s}}^{T}}{s},
\end{equation}
Differentiating ${{V}_{2}}$ with respect to time, one obtains
\begin{equation}\label{eq25}
{{\dot{V}}_{2}}={{\dot{V}}_{1}}+{{{s}}^{T}}{\dot{s}}
\end{equation}
In order to guarantee the sliding surface to ultimately converge to zero, the control scheme should include two sub-laws. The first is the switching law, which is employed to drive the system states towards a particular sliding surface. This switching control signal is given by
\begin{equation}\label{eq27}
{{{u}}_{sw}}=-M\left( {{c}_{2}}sign\left( {{s}} \right)+{{c}_{3}}{s} \right),
\end{equation}
$c_2$ and $c_3$ are the positive definite diagonal matrices.
The second is the equivalent control law that is utilized to keep those states lying on the sliding surface. The equivalent control signal is formulated by
\begin{equation}\label{eq26}
 {{{u}}_{eq}}=M{{{{\dot{\alpha }}}}_{2f}}-K-M{{\lambda }}{{{\dot{{z}}}}_{1}}=-M\left( {{\lambda }}{{{\dot{{z}}}}_{1}}+{{M}^{-1}}K-{{{{\dot{\alpha }}}}_{2f}} \right)  
\end{equation}
Therefore, the total control signal can be formed by
\begin{equation}\label{eq28}
\begin{array}{r@{}l@{\qquad}l}
{u}=&{{{u}}_{sw}}+{{{u}}_{eq}} \\
=&-M\left( {{c}_{2}}sign\left( {{s}} \right)+{{c}_{3}}{s} \right)-M\left( {{\lambda }}{{{\dot{{z}}}}_{1}}+{{M}^{-1}}K-{{{{\dot{\alpha }}}}_{2f}} \right) .
\end{array}
\end{equation}
Stability of the proposed control scheme in (\ref{eq28}) is analysed in the following theorem.

\textbf{Theorem 1. }
The proposed control law (\ref{eq28}) guarantees the closed-loop system (\ref{eq14}) to be asymptotically stable.

\begin{proof}
From (\ref{eq25}), the derivative of the second Lyapunov function candidate, ${{\dot{V}}_{2}}$, can be rewritten by
\begin{equation}\label{eq29}
\begin{array}{r@{}l@{\qquad}l}
 {{{\dot{V}}}_{2}}=&-{{{{z}}}_{1}}^{T}{{c}_{1}}{{{{z}}}_{1}}+{{{{s}}}^{T}}\left( {{\lambda }}{{{\dot{{z}}}}_{1}}+{{M}^{-1}}\left( K+{u}-M{{{{\dot{\alpha }}}}_{2f}} \right) \right) \\ 
 =&-{{{{z}}}_{1}}^{T}{{c}_{1}}{{{{z}}}_{1}}-{{s}}^{T}{{c}_{2}}sign({s})-{{s}}^{T}{{c}_{3}}{s} \\
 &+{{{s}}^{T}}\left( {{c}_{2}}sign\left( s \right)+{{\lambda }}{{{\dot{{z}}}}_{1}}+{{M}^{-1}}\left( K+{u}-M{{{{\dot{\alpha }}}}_{2f}} \right)+{{c}_{3}}{s} \right).  
\end{array}
\end{equation}
Substituting the control input in (\ref{eq28}) into (\ref{eq29}) leads to
\begin{equation}\label{30}
{{\dot{V}}_{2}}=-{{{z}}_{1}}^{T}{{c}_{1}}{{{z}}_{1}}-{{s}}^{T}{{c}_{2}}sign\left( {{s}} \right)-{{s}^{T}}{{c}_{3}}{s}<0.
\end{equation}
Therefore, based on the Lyapunov stability theory the sliding surface $s$ is asymptotically stable. 
\end{proof}
	
\subsection{Adaptive dynamic surface control for uncertain DAR systems}
The deterministic control law (\ref{eq28}) can be effectively employed provided that the system parameters are certain. Nevertheless, in practice, the DAR system operates under system uncertainties such as parameter variations and nonlinearities or external disturbances. In other words, it is impractical to accurately determine the system parameters in $K(\theta ,\dot{\theta },\ddot{\theta })$. To deal
with these challenges, it is proposed to utilize the radial basis function neural network (RBFN) to approximately estimate the dynamic model $K(\theta ,\dot{\theta },\ddot{\theta })$, given the system uncertainties.

\begin{figure}[htbp]
	\centering
	\includegraphics[width=0.8\linewidth]{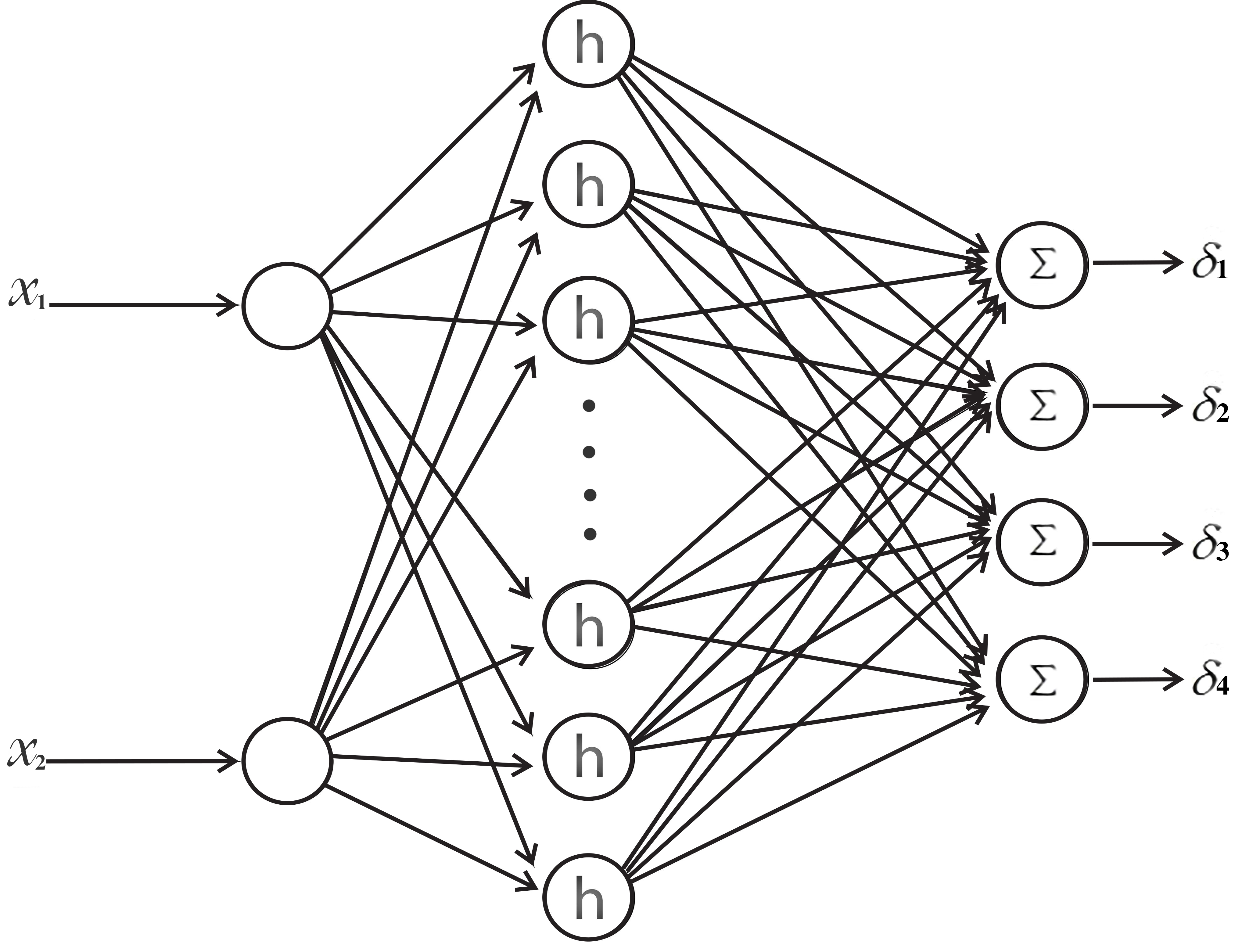}
	\caption{Schematic diagram of RBF neural network.}
	\label{fig4}
\end{figure}

Overall, the structure of the RBFN \cite{RN27} as shown in Fig. \ref{fig4} comprises the inputs $(x_1,x_2)$, the outputs $(\delta_1,\delta_2,\delta_3,\delta_4)$ and a number of neurons in the hidden layers. If ${r}={{\left( {{{{x}}}_{1}}^{T},{{{{x}}}_{2}}^{T} \right)}^{T}}$ and $\delta=(\delta_1,\delta_2,\delta_3,\delta_4)^T$, then the output of the RBFN can be presented by
\begin{equation}
	\delta\left( r \right)={{W}^{T}}h\left( r \right),
	\end{equation}
where $W$ is the weight matrix, $h\left( r \right)={{({{h}_{1}}\left( r \right),{{h}_{2}}\left( r \right),...,{{h}_{l}}\left( r \right) })^{T}}$, where ${{h}_{i}}\left( r \right)$ is an activation function. The widely used activation function, which is also employed in this work, is Gaussian,
\begin{equation}\label{eq33}
{{h}_{i}}(r)=\frac{\exp \left( \frac{{{\left\| {{x}_{1}}-{{\rho }_{1i}} \right\|}^{2}}+{{\left\| {{x}_{2}}-{{\rho}_{2i}} \right\|}^{2}}}{{{b}_{i}}^{2}} \right)}{\sum\limits_{j=1}^{n}{\exp \left( -\frac{{{\left\| {{x}_{1}}-{{\rho}_{1j}} \right\|}^{2}}+{{\left\| {{x}_{2}}-{{\rho}_{2j}} \right\|}^{2}}}{{{b}_{j}}^{2}} \right)}}, \;\; i=1,2,...,n,
\end{equation}
where $n$ is the number of neurons in the hidden layer, $\rho$ is the matrix of means and $b$ is the vector of variances.

If $\hat{W}$ denotes estimation of the weight matrix $W$, which is updated by the adaptation mechanism as follows, 
\begin{equation}\label{eq34}
\dot{\hat{W}}=\Gamma \left( {h}{{{{s}}}^{T}}-\varsigma \left\| {{s}} \right\|\hat{W} \right)
\end{equation}
where $\varsigma$ is positive and $\Gamma$ is the positive definite diagonal matrix of the adaptation constants, then the output of the RBFN $\delta(r)$ is approximated by
\begin{equation}
{\hat{\delta }}\left( r \right)={{\hat{W}}^{T}}{h}.
\end{equation}
In this work, we employ the RBFN to adaptively estimate the uncertain dynamic $K$; therefore, the control input in (\ref{eq28}) can be approximated by
\begin{equation}\label{eq35}
{u}=-M\left( {{c}_{2}}sign\left( {{s}} \right)+{{c}_{3}}{s} \right)-M\left( {{\lambda }}{{{\dot{{z}}}}_{1}}+{{M}^{-1}}{{\hat{W}}^{T}}{h}-{{{{\dot{\alpha }}}}_{2f}} \right) .
\end{equation}
 
It is noticed that the estimation of the weight matrix in the RBFN is derived from the Lyapunov theory, which guarantees stability of the closed-loop system as presenting in the following theorem.

\textbf{Theorem 2. }
Given the adaptation mechanism (\ref{eq34}), the proposed control scheme (\ref{eq35}) can guarantee the closed-loop DAR system (\ref{eq15}) to be input-to-state stable \cite{RN28} with the attractor
\begin{equation}\label{eq36}
D=\left\{ {s}\in {{R}^{4}}|\left\| {s} \right\|> \frac{{{\varepsilon }_{N}}+\varsigma \frac{{{\left\| W \right\|}_{F}}^{2}}{4}}{{{c}_{3\min }}} \right\},
\end{equation}
where ${c}_{3\min }$ is the minimum value of $c_3$, and ${\varepsilon }_{N}$ is a small positive number so that the approximation error $\varepsilon=\delta-\hat{\delta}$ satisfies $\left\| {\varepsilon}  \right\|<{\varepsilon }_{N}$.

\begin{proof}
Let
\begin{equation}
\tilde{W}=W-\hat{W}
\end{equation}
define the error between the ideal weight $W$ and the estimated weight $\hat{W}$
Considering the Lyapunov function candidate
\begin{equation}\label{eq37}
{{V}_{2}}={{V}_{1}}+\frac{1}{2}{{{s}}^{T}}{s}+tr\left( {{{\tilde{W}}}^{T}}{{\Gamma }^{-1}}\tilde{W} \right)
\end{equation}
and differentiating it with respect to time, one obtains
\begin{equation}\label{eq38}
\begin{array}{r@{}l@{\qquad}l}
{{{\dot{V}}}_{2}}=&{{{\dot{V}}}_{1}}+{{{{s}}}^{T}}\dot{{s}}+tr\left( {{{\tilde{W}}}^{T}}{{\Gamma }^{-1}}\dot{\tilde{W}} \right) \\ 
 =&-{{{{z}}}_{1}}^{T}{{c}_{1}}{{{{z}}}_{1}}-{{{{s}}}^{T}}{{c}_{2}}sign\left( {{s}} \right)-{{{{s}}}^{T}}{{c}_{3}}{s}+{{{{s}}}^{T}}\left( {\delta }-{\hat{\delta }} \right) \\
 &-tr\left( {{{\tilde{W}}}^{T}}{{\Gamma }^{-1}}\dot{\hat{W}} \right) \\ 
=&-{{{{z}}}_{1}}^{T}{{c}_{1}}{{{{z}}}_{1}}-{{{{s}}}^{T}}{{c}_{2}}sign\left( {{s}} \right)-{{{{s}}}^{T}}{{c}_{3}}{s}+{{{{s}}}^{T}}{{W}^{T}}{h}\\
&-{{{{s}}}^{T}}{{{\hat{W}}}^{T}}{h}-tr\left( {{{\tilde{W}}}^{T}}{{\Gamma }^{-1}}\dot{\hat{W}} \right)  +{{{{s}}}^{T}}{\varepsilon}  \\ 
=&-{{{{z}}}_{1}}^{T}{{c}_{1}}{{{{z}}}_{1}}-{{{{s}}}^{T}}{{c}_{2}}sign\left( {{s}} \right)-{{{{s}}}^{T}}{{c}_{3}}{s}+{{{{s}}}^{T}}{\varepsilon}\\
&+tr\left( {{{\tilde{W}}}^{T}}\left( {{{h}{{{s}}}^{T}-\Gamma }^{-1}}\dot{\hat{W}}\right) \right).  
\end{array}
\end{equation}
If we substitute (\ref{eq35}) into the derivative of the Lyapunov function (\ref{eq38}), it yields
\begin{align}\label{eq39}
{{\dot{V}}_{2}}=&-{{{z}}_{1}}^{T}{{c}_{1}}{{{z}}_{1}}-{{{s}}^{T}}{{c}_{2}}sign\left( {s} \right)-{{{s}}^{T}}{{c}_{3}}{s}+{{{s}}^{T}}{\varepsilon}\\ \nonumber
&+\varsigma \left\| {{s}} \right\|tr\left( {{{\tilde{W}}}^{T}}\left( W-\tilde{W} \right) \right).
\end{align}
By the use of Cauchy-Schwarz inequality
\begin{equation}\label{eq40}
tr\left[ {{{\tilde{W}}}^{T}}\left( W-\tilde{W} \right) \right]\le {{\left\| {\tilde{W}} \right\|}_{F}}{{\left\| W \right\|}_{F}}-{{\left\| {\tilde{W}} \right\|}_{F}}^{2},
\end{equation}
we can easily compute the inequality of the derivative of the Lyapunov function as follows,
\begin{align}\label{eq41}
{{\dot{V}}_{2}}\le &-{{{z}}_{1}}^{T}{{c}_{1}}{{{z}}_{1}}-{{{s}}^{T}}{{c}_{2}}sign\left( {s} \right)-{{{s}}^{T}}{{c}_{3}}{s}+{{{s}}^{T}}{\varepsilon}\\ \nonumber
&+\varsigma \left\| {{s}} \right\|\left( {{\left\| {\tilde{W}} \right\|}_{F}}{{\left\| W \right\|}_{F}}-{{\left\| {\tilde{W}} \right\|}_{F}}^{2} \right).
\end{align}
Rearranging (\ref{eq41}) by utilizing the attractor (\ref{eq36}), it yields
\begin{equation}\label{eq42}
\begin{array}{r@{}l@{\qquad}l}
{{\dot{V}}_{2}}\le &-{{{s}}^{T}}{{c}_{2}}sign\left( {{s}} \right)-\varsigma \left\| {{s}} \right\|{{\left( {{\left\| {\tilde{W}} \right\|}_{F}}-\frac{1}{2}{{\left\| W \right\|}_{F}} \right)}^{2}}\\
&-\left\| {{s}} \right\|{{c}_{3\min }}\left\| {{s}} \right\|+{{\varepsilon }_{N}}\left\| {{s}} \right\|+\left\| {{s}} \right\|\varsigma \left( \frac{{{\left\| W \right\|}_{F}}^{2}}{4} \right)
\end{array}
\end{equation}
In other words, if the sliding surface is outside the attractor, which is 
\begin{equation}
\left\| {s} \right\|>\frac{{{\varepsilon }_{N}}+\varsigma \frac{{{\left\| W \right\|}_{F}}^{2}}{4}}{{{c}_{3\min }}},
\end{equation}
we then have ${{\dot{V}}_{2}}\le 0$. Therefore, the sliding surface ${s}$ is input-to-state stable. 	
\end{proof}	
	
\section{Simulation Results}
\label{sec_4}
To demonstrate effectiveness of the proposed control law, we conducted experiments in simulation environment. To simulate the DAR protocol, the two robot arms were first to track the desired trajectories to reach the object. The reference trajectories in the first 2 seconds are mathematically specified by
\begin{equation}\label{eq45}
\begin{array}{r@{}l@{\qquad}l}
	{{x}_{a1}}(t)&={{x}_{f1}}+({{x}_{i1}}-{{x}_{f1}}){{e}^{-10{{t}^{2}}}},  \\
	
	{{y}_{a1}}(t)&={{y}_{f1}}+({{y}_{i1}}-{{y}_{f1}}){{e}^{-10{{t}^{2}}}},  \\
	
	{{x}_{a2}}(t)&={{x}_{f2}}+({{x}_{i2}}-{{x}_{f2}}){{e}^{-10{{t}^{2}}}},   \\
	
	{{y}_{a2}}(t)&={{y}_{f2}}+({{y}_{i2}}-{{y}_{f2}}){{e}^{-10{{t}^{2}}}},  
	\end{array}
	\end{equation}
where $x_{a1},y_{a1},x_{a2},y_{a2}$ are the trajectories of the robot arms.  $\left( {{x}_{i1}},{{y}_{i1}},{{x}_{i2}},{{y}_{i2}} \right)$ and $\left( {{x}_{f1}},{{y}_{f1}},{{x}_{f2}},{{y}_{f2}} \right)$ are the initial and final positions of the manipulators, respectively. After firmly holding the payload, the robot transports the object along the half of a circle so that it can avoid collision with an obstacle. The center of the object is expected to travel on a curve as follows,
	\begin{equation}
	\begin{array}{r@{}l@{\qquad}l}
	{{x}_{mr}}(t)&={{x}_{0}}+{{r}_{m}}\cos(\phi t), \\
	
	{{y}_{mr}}(t)&={{y}_{0}}+{{r}_{m}}sin(\phi t),
	\end{array}
	\end{equation}
where $\left( {{x}_{0}},{{y}_{0}} \right)$ is the position of the obstacle, which is also the center of the circle on which the center of the object travels. ${{r}_{m}}$ is the radius of the circle, while $\phi $ is a polar angle that varies from $-\pi$ to $0$. It is noted that the joint angles between the link and the base or its preceding link at the beginning $t=0$ were known, ${{q}_{1}}(0)=\frac{\pi }{6},\,\,{{q}_{2}}(0)=\frac{\pi }{2},\,\,{{q}_{3}}(0)=\pi$ and ${{q}_{4}}(0)=\frac{-2\pi }{3}$.

\begin{figure}[htbp]
		\centering
		\includegraphics[width=0.8\linewidth]{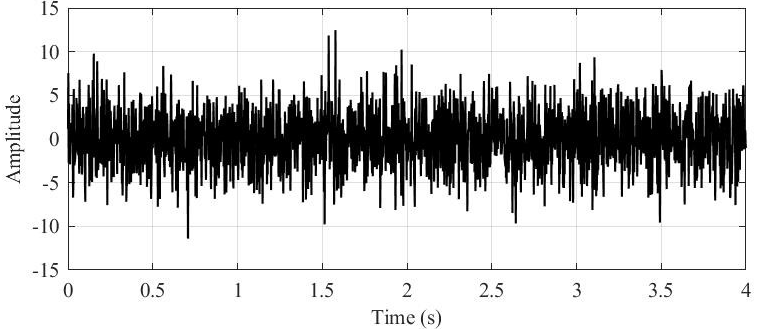}
		\caption{External disturbance}
		\label{fig5}
	\end{figure}

In the synthetic experiments, the physical model parameters of the DAR system were given. Furthermore, the parameters of the DSC controller were known. Those information are summarized in Table \ref{tab1}. It was supposed that there is no prior knowledge of the robot dynamics, then the weight matrix $W$ of the RBFN were initialized by zeros. Moreover, an unexpected disturbance as shown in Fig. \ref{fig5}, which exerts the applied forces, was taken into consideration to illustrate robustness of the proposed approach.
	
\begin{table}
	\caption{Parameters of the dual arm robot system} 
	\begin{center}
	\label{tab1} 
	\begin{tabular}{|c|} 
		\hline
		Dynamic model parameters\\
		\hline
		${{m}_{1}}={{m}_{2}}={{m}_{3}}={{m}_{4}}=1.5\,(kg)$;\\  ${{I}_{1}}={{I}_{2}}={{I}_{3}}={{I}_{4}}=0.18\,(kg{{m}^{2}})$;\\
		${{l}_{1}}={{l}_{2}}={{l}_{3}}={{l}_{4}}=1.2\,(m)$;\\  ${{k}_{1}}={{k}_{2}}={{k}_{3}}={{k}_{4}}=0.48\,(m)$;\\  
		${{b}_{1}}={{b}_{2}}={{b}_{3}}={{b}_{4}}=110\,(Nm/s)$;\\  
		${{d}_{1}}=0.25\,(m)$; ${{d}_{2}}=1.2\,(m)$; $\mu =0.35$; $m=1.5\,(kg)$\\ 
		\hline
		Reference trajectory parameters\\
		\hline
		$\left( {{x}_{i1}},{{y}_{i1}},{{x}_{i2}},{{y}_{i2}} \right)=\left( 0.76,\,0.6,\,-0.76,\,0.6 \right)$;\\
		$\left( {{x}_{f1}},{{y}_{f1}},{{x}_{f2}},{{y}_{f2}} \right)=\left( -0.275,\,1.4,\,-0.525,\,1.4 \right)$;\\
		$\left( {{x}_{0}},{{y}_{0}} \right)=\left( 0,\,1.4 \right)$;  ${{r}_{m}}=0.4$;\\
		${{\theta}_{1}}(0)=\frac{\pi }{6};\,\,{{\theta}_{2}}(0)=\frac{\pi }{2};\,\,{{\theta}_{3}}(0)=\pi ;\,\,{{\theta}_{4}}(0)=\frac{-2\pi }{3}$;\\
		${{\dot{\theta}}_{1}}(0)={{\dot{\theta}}_{2}}(0)={{\dot{\theta}}_{3}}(0)={{\dot{\theta}}_{4}}(0)=0$\\
		\hline
		Controller parameters\\
		\hline
		$\lambda =diag\left( 15,15,15,15 \right)$; \\
		${{c}_{1}}=diag\left( 122,122,122,122 \right)$;\\
		${{c}_{2}}=diag\left( 122,122,122,122 \right)$;\\ 
		${{c}_{3}}=diag\left( 152,152,152,152 \right)$;\\ 
		$\hat{W}\left( 0 \right)=0$;  $\Gamma =diag\left( 30,30,30,30 \right)$\\
		\hline
	\end{tabular}
	\end{center}
\end{table}

\begin{figure*}[htbp]
		\centering
		\subfloat[]{\label{fig6a}
		\includegraphics[width=0.48\linewidth]{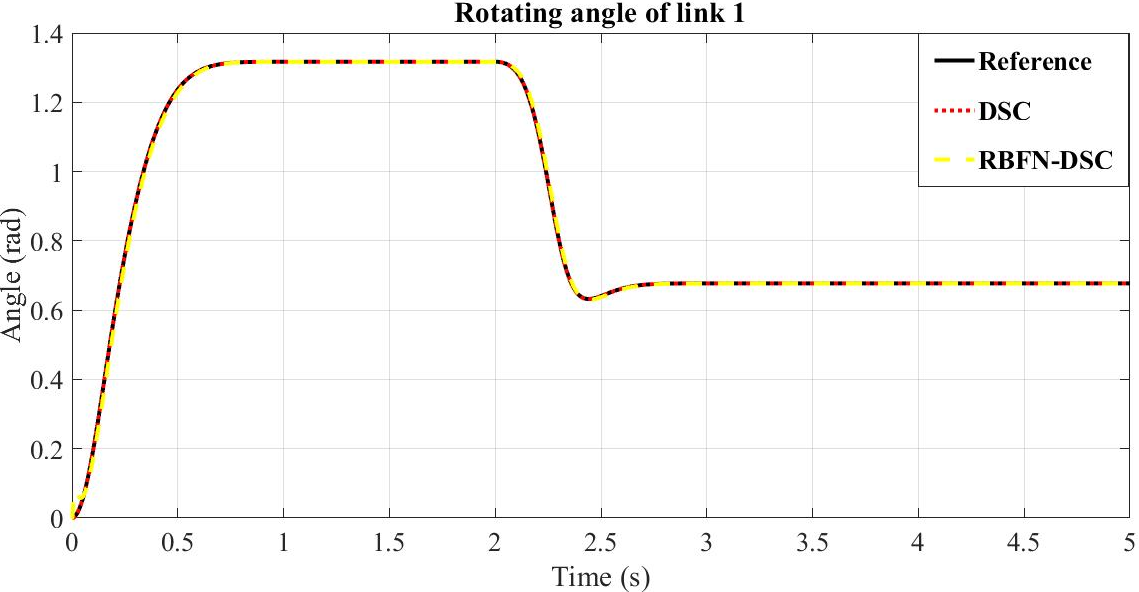}} 
		\subfloat[]{\label{fig6b}	
		\includegraphics[width=0.48\linewidth]{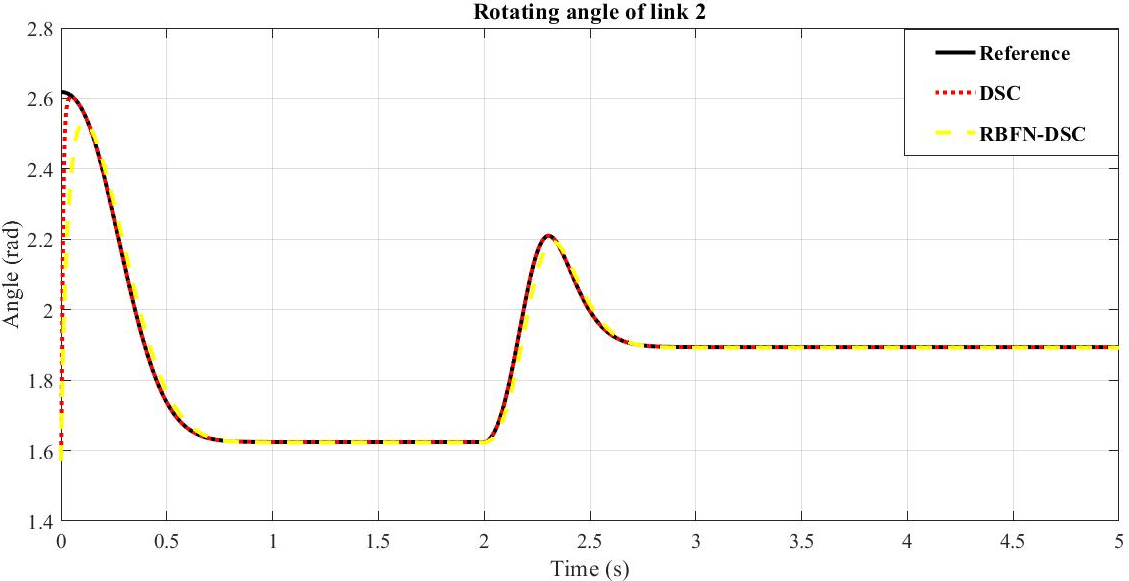}} \\
%
		\subfloat[]{\label{fig6c}
		\includegraphics[width=0.48\linewidth]{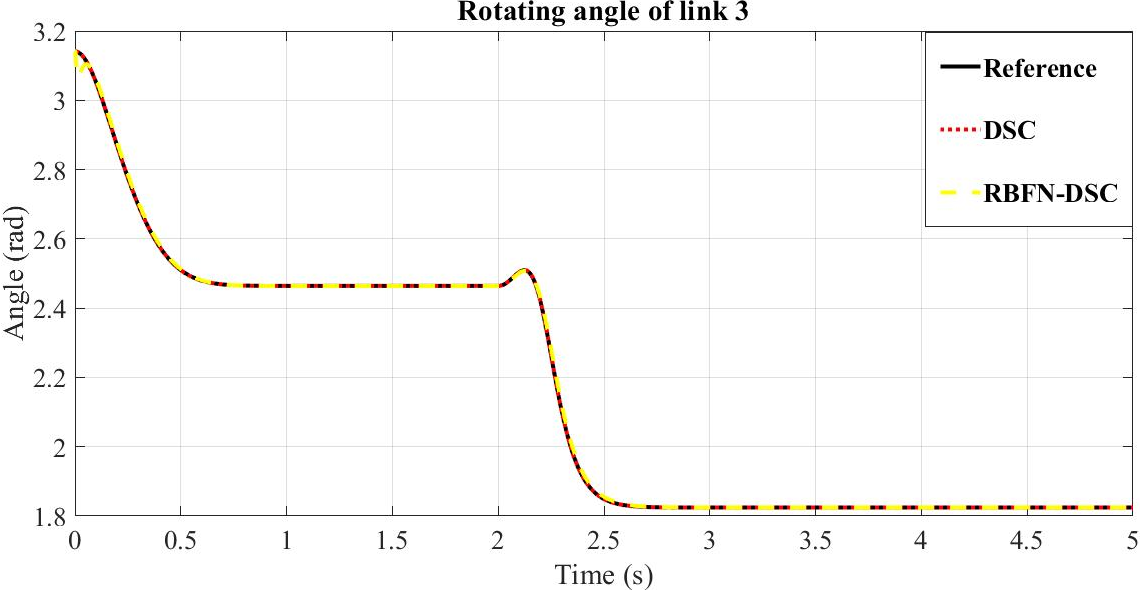}} 
		\subfloat[]{\label{fig6d}	
		\includegraphics[width=0.48\linewidth]{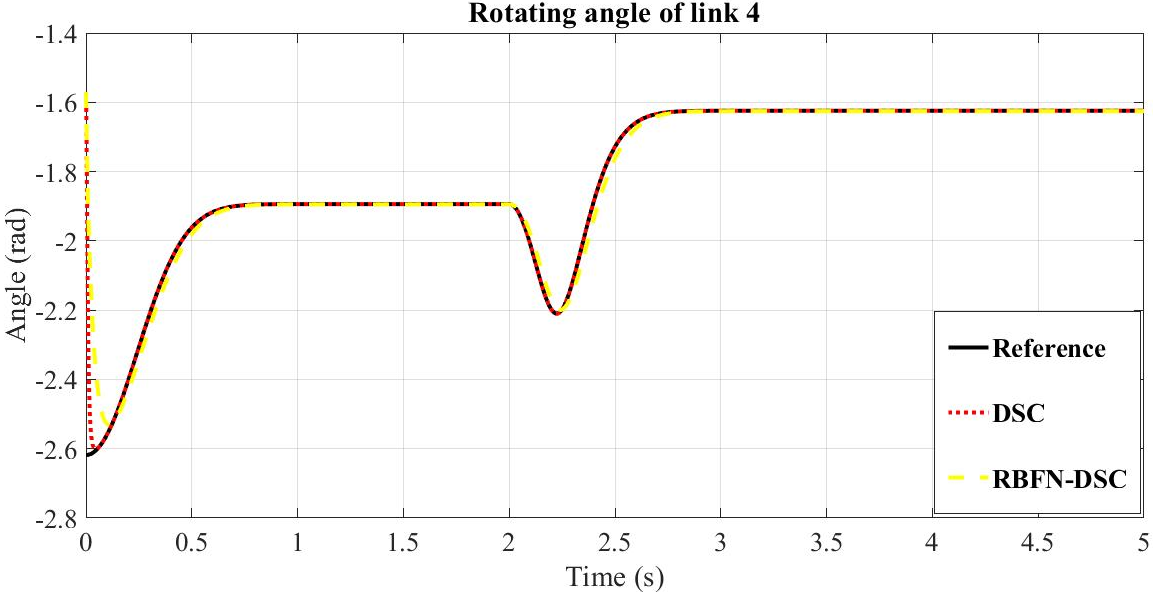}} \\
	\caption{Joint angles of the link and the base or its preceding link: (a) first link, (b) second link, (c) third link and (d) fourth link.}
	\label{fig6}
		\end{figure*}

\begin{figure*}[htbp]
		\centering
		\subfloat[]{\label{fig7a}
		\includegraphics[width=0.48\linewidth]{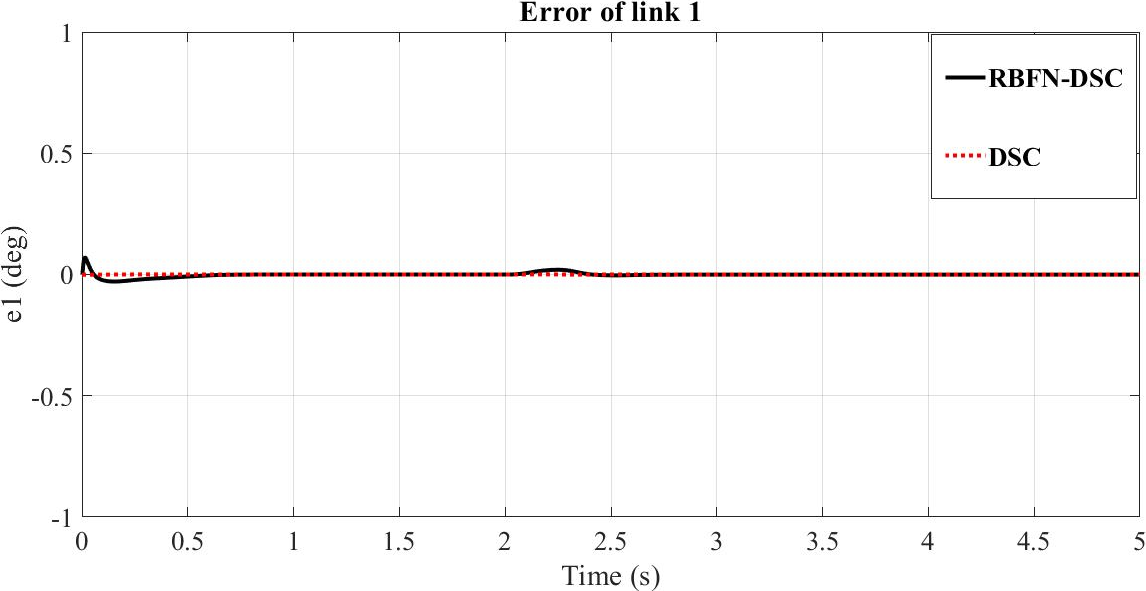}} 
		\subfloat[]{\label{fig7b}	
		\includegraphics[width=0.48\linewidth]{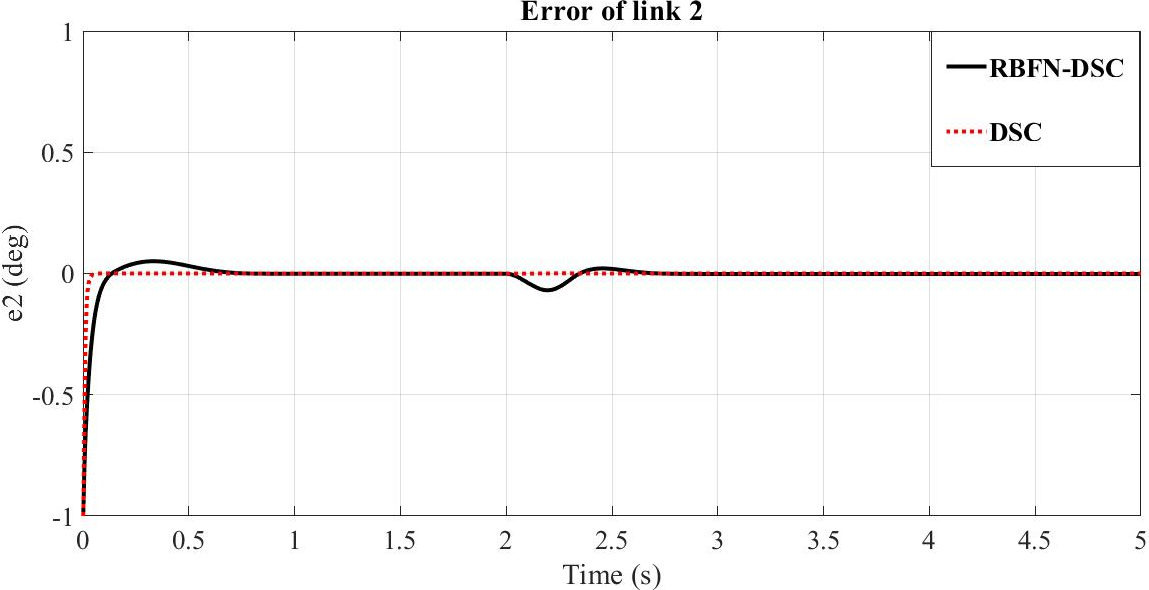}} \\
%
		\subfloat[]{\label{fig7c}
		\includegraphics[width=0.48\linewidth]{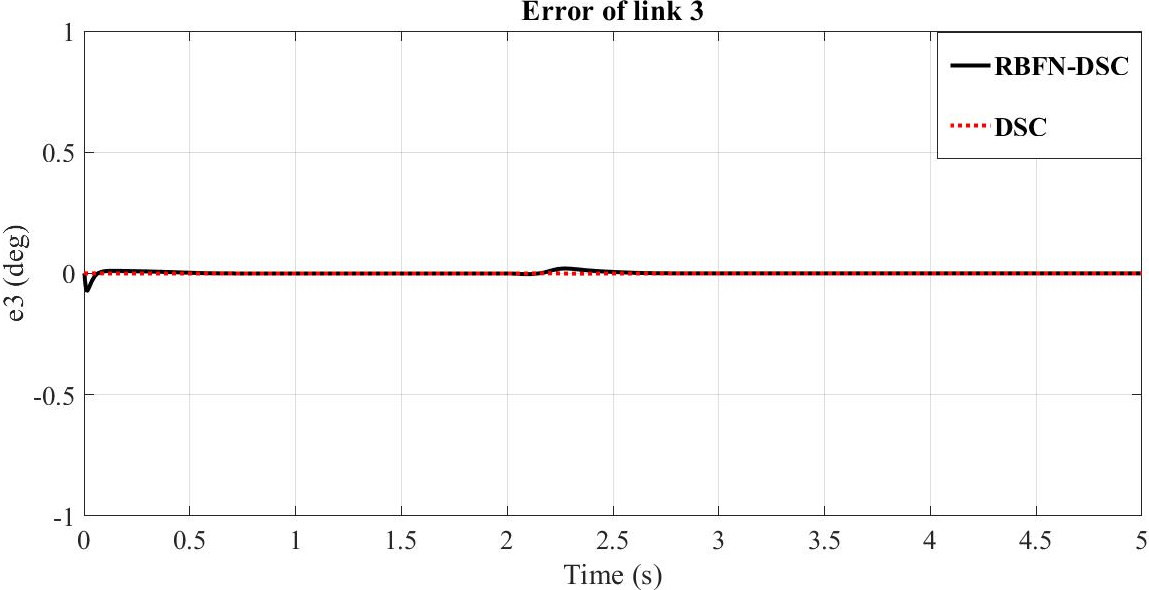}} 
		\subfloat[]{\label{fig7d}	
		\includegraphics[width=0.48\linewidth]{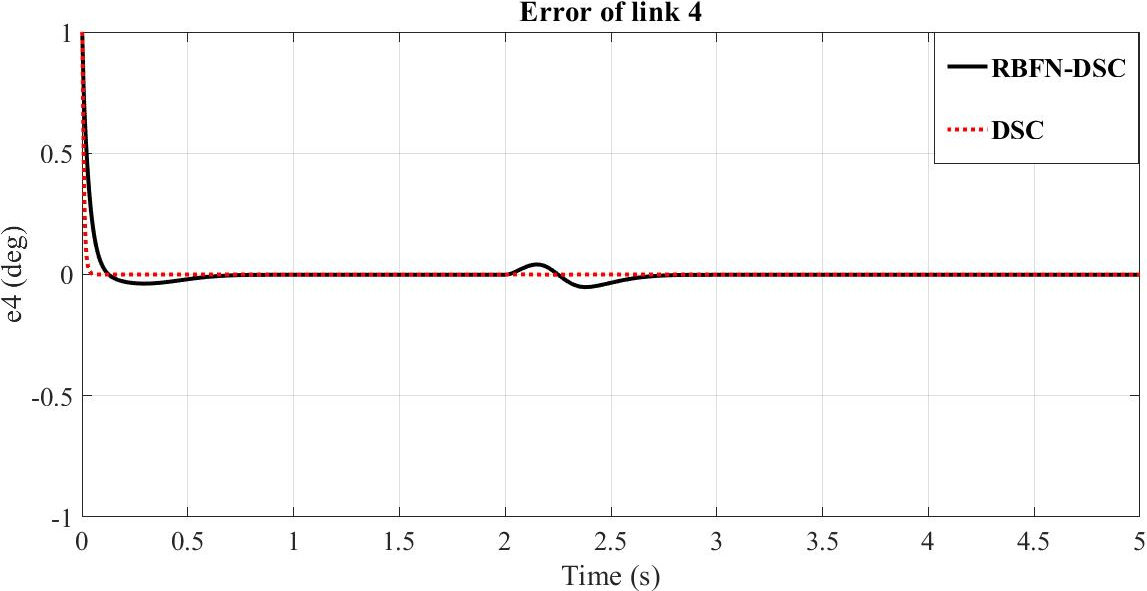}} \\
	\caption{Errors of joint angles of the link and the base or its preceding link: (a) first link, (b) second link, (c) third link and (d) fourth link.}
	\label{fig7}
		\end{figure*}

Before examining the motions of the robot arms, let's investigate the motions of the four links of the DAR system by considering the joint angles between the link and the base or its preceding link on $xy-plane$ as illustrated in Fig. \ref{fig5}. For the purposes of comparisons, in this experimental example we implemented both the algorithms of the conventional DSC scheme as presented in Section \ref{sec_3a}, where the system parameters were assumed to be determined and certain, and the proposed RBFN based DSC (RBFN-DSC) approach. It can be clearly seen in Fig. \ref{fig6} that the results obtained by the two implemented algorithms were expected to approach the references, which are early obtained from the equations in (\ref{eq45}), all the time. While the deterministic DSC control law quickly tracked the references in all the links, the proposed algorithm performed well in the lower links as shown in Figures \ref{fig6a} and \ref{fig6c} and insignificantly degraded in the upper links as shown in Figures \ref{fig6b} and \ref{fig6d}, approximate 0.05 s behind the DSC method. This is understandable since the parameters of the DSC control law were given, while the RBFN-DSC technique needed time to adaptively estimate those. Nonetheless, in fact, given the system uncertainties including the system parameter variations, actuator nonlinearities and external disturbances, exactly determining the system parameters for the DSC algorithm is not really practical, while the proposed approach can estimate those parameters by the use of the RBFN. Errors of the joint angles at the links as illustrated in Fig. \ref{fig7} consolidate efficacy of the proposed control scheme.
		
\begin{figure*}[htbp]
		\centering
		\subfloat[]{\label{fig8a}
		\includegraphics[width=0.48\linewidth]{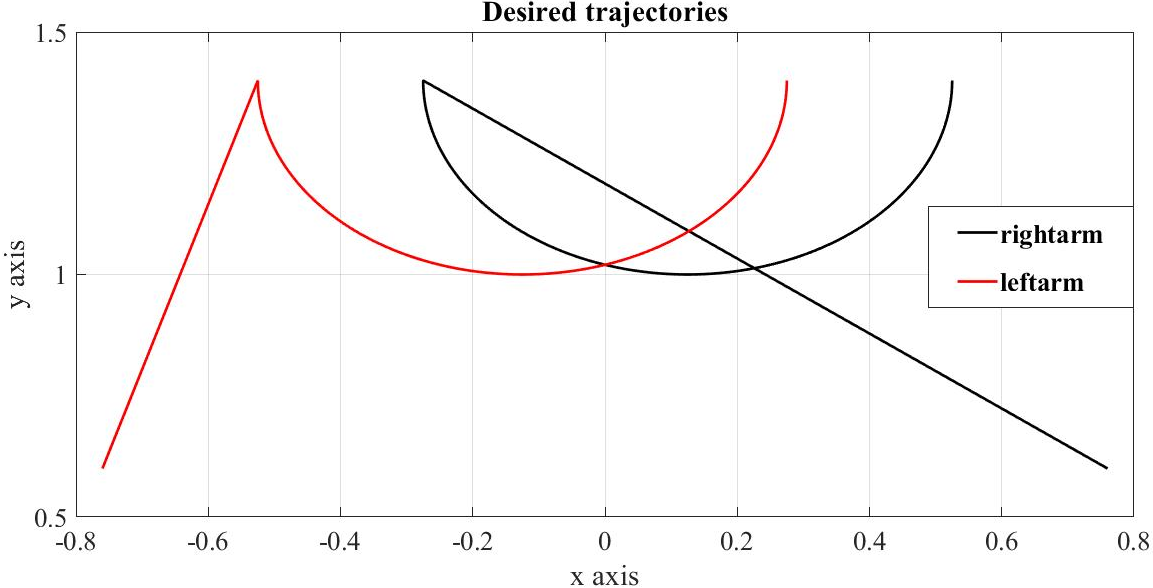}}  \\
%
		\subfloat[]{\label{fig8b}
		\includegraphics[width=0.48\linewidth]{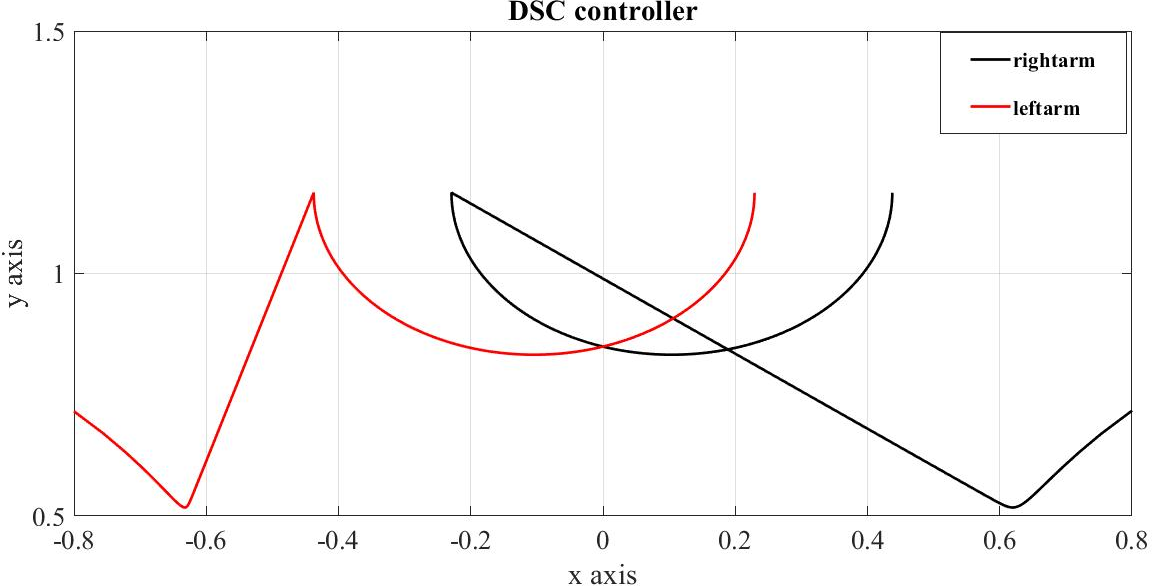}} 
		\subfloat[]{\label{fig8c}	
		\includegraphics[width=0.48\linewidth]{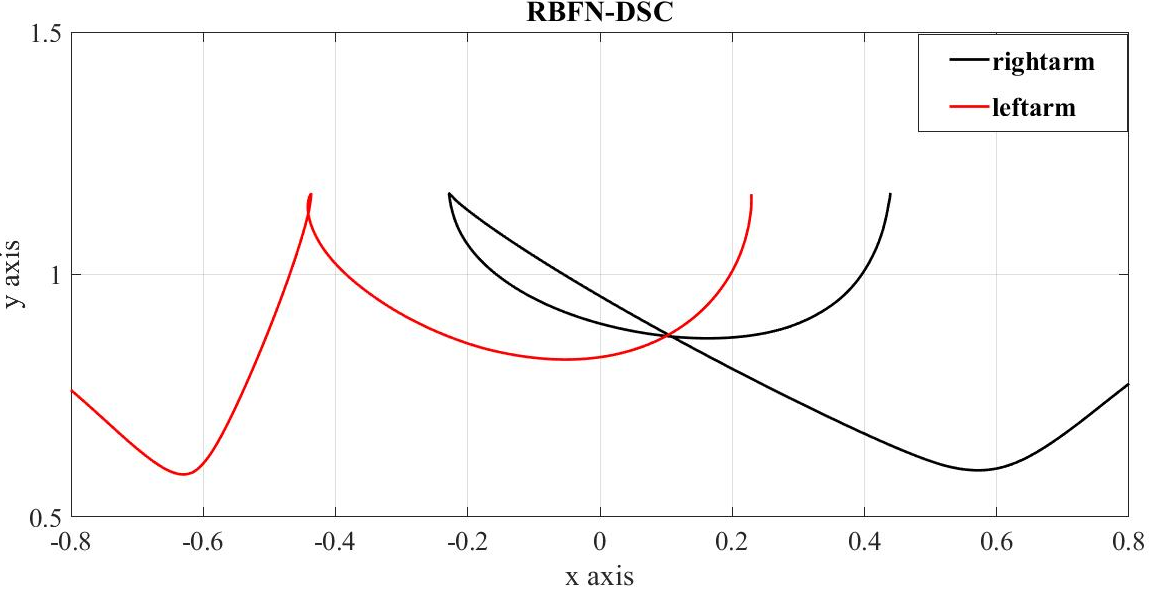}} \\
	\caption{Motion trajectories of the end-effectors: (a) Expected trajectories, (b) trajectories obtained by the DSC and (c) trajectories obtained by the RBFN-DSC.}
	\label{fig8}
		\end{figure*}

More importantly, as can be seen in Fig. \ref{fig8}, the motion trajectories of the two end-effectors show that the proposed RBFN-DSC is really efficiently and effectively practical. Given the aim of transporting the payload along a half of a circle to avoid collision with an obstacle, the movements of both the left and right manipulators of the robot under the control of the deterministic DSC scheme in Fig. \ref{fig8b} and the proposed RBFN-DSC law in Fig. \ref{fig8c} were expected to track the ideal trajectories as shown in Fig. \ref{fig8a}. It can be clearly seen that given the system parameters, the DSC method controlled the end-effectors to travel quite smoothly, including before approaching the payload and during transporting it, as compared with the desired references. Nevertheless, though having to estimate the system parameters under their uncertainties and nonlinearities, the trajectories obtained by the proposed RBFN-DSC control scheme in the whole protocol are highly comparable to not only those obtained by the DSC method but also the expectation. That is, the proposed algorithm guarantees that the DAR system to be able to adaptively learn its nonlinear parameters while safely transport the payload to the destination. The RBFN-DSC control law is highly applicable for the uncertain DAR systems.
	
\section{Conclusions}
\label{sec_5}
The paper has introduced a new robust adaptive control approach for an uncertain DAR system, where the manipulators are expected to grasp and transport an object to a destination on the desired trajectories. To guarantee motions of the robot's end-effectors to be robustly tracked on the references, the control scheme is designed based on the DSC technique. Nonetheless, due to the system uncertainties and nonlinearities, the DAR system dynamics are not practically determined, which leads to impracticality of the DSC algorithm. Hence, it has been proposed to adaptively learn the uncertain system parameters by employing the RBFN, where the adaptation mechanism has been derived from the Lyapunov function to guarantee the stability of the closed-loop control system. The results obtained by a synthetic implementation have verified the proposed control law. It is noted that the proposed algorithm will be implemented in the realistic DAR system in the future works.

\bibliographystyle{unsrt}      
\bibliography{References}   


\end{document}